\documentclass{article}

\usepackage{arxiv}
\usepackage{longtable}
\usepackage{array}
\usepackage{cite}
\usepackage{amsmath}
\usepackage{mathtools}

\usepackage[utf8]{inputenc} 
\usepackage[T1]{fontenc}    
\usepackage{hyperref}       
\usepackage{url}            
\usepackage{booktabs}       
\usepackage{amsfonts}       
\usepackage{nicefrac}       
\usepackage{microtype}      
\usepackage{lipsum}
\usepackage{graphicx}
\graphicspath{ {./images/} }
\usepackage{subcaption}

\title{Dominant motion identification of multi-particle system using deep learning from video.}

\author{
 Yayati Jadhav \\
  Mechanical Engineering\\
  Carnegie Mellon University\\
  Pittsburgh, PA 15213 \\
  \texttt{yayatij@andrew.cmu.edu} \\
  \And Amir Barati Farimani
  \\
  Mechanical Engineering\\
  Carnegie Mellon University\\
  Pittsburgh, PA 15213 \\
  \texttt{barati@cmu.edu} \\
}

\begin{document}
\maketitle
\begin{abstract}
Identifying underlying governing equations and physical relevant information from high-dimensional observable data has always been a challenge in physical sciences. With the recent advances in sensing technology and available datasets, various machine learning techniques have made it possible to distill underlying mathematical models from sufficiently clean and usable datasets. However, most of these techniques rely on prior knowledge of the system and noise free data obtained by simulation of physical system or by direct measurements of the signals. Hence, the inference obtained by using these techniques is often unreliable to be used in the real world where observed data is noisy and requires feature engineering to extract relevant features. In this work, we provide a deep-learning framework that extracts relevant information from real-world videos of highly stochastic systems, with no prior knowledge and distills the underlying governing equation representing the system. We demonstrate this approach on videos of confined multi-agent/particle systems of ants, termites, fishes as well as a simulated confined multi-particle system with elastic collision interactions. Furthermore, we explore how these seemingly diverse systems have predictable underlying behavior.  In this study, we have used computer vision and motion tracking to extract spatial trajectories of individual agents/particles in a system, and by using LSTM VAE we projected these features on a low-dimensional latent space from which the underlying differential equation representing the data was extracted using SINDy framework.
\end{abstract}

\section{Introduction}

Learning specific action cases from one systems and applying it to a different systems is one of the major research areas in machine learning. Nature provides a variety of well optimized action case scenarios, mimicking these natural systems and applying it to the other systems is a well researched topic in biological inspired computation. Distilling underlying dynamics of a system from data is a form of data-driven pattern recognition where similarities between the systems states over time are found. Recognition of this pattern not only helps in prediction of future states but it also helps to identify and quantify the deviation of a systems from ideal state due to disturbance. One of the major challenge in distilling underlying dynamics from data is the availability of specific features which define the states of the system. This problem is exacerbated in case of video data, where the data features are observable spatio-temporal data and required feature engineering and prior knowledge of the system to find a surrogate system model. Visually learning the dynamics of a system, say a group of ants foraging in space,  in the form of a differential equation from videos of the systems not only allows for understanding of the system better but can also be used to transfer dynamics of this system to another system like swarm robots or optimization algorithms.  Recent advancements in machine learning techniques has made it possible to approximate representative mathematical models directly from observable data \cite{1,2,3} instead of building models from first principle physical laws.

Time-series data describing evolution of states of a system over time, is one of the most abundant form of data available. Underlining dynamics of a nonlinear dynamical system can be extracted from time series data using symbolic regression \cite{3,11}. A more popular uses sparse regression, as demonstrated in SINDy framework, a predefined candidate library of all possible data features is created and spare regression is used to select a parsimonious model that represents most prominent dynamics of the system\cite{10,45,46}.  However, since numerical approximation of derivatives from data is required for sparse regression, this approach is highly susceptible to noise. Moreover, this approach fails to handle spatio-temporal data where x and y co-ordinates from 1 feature\cite{31}
and becomes trickier with the increase in number of features/variables observed in the system, as it becomes difficult to objectively eliminate features and their interactions based on their contribution to the system dynamics. An extension of SINDy framework where auto encoder is used to find effective coordinate system representing the system in fewest possible feature\cite{32}
, solves this dimentionality problem to some extent, however relatively clean and noise free data is still required for this approach. Besides sparse regression based methods, techniques such as Gaussian process is also used to find the governing equation from the data. In this process, a Gaussian prior is placed on an unknown coefficients which is inferred using maximum likely hood estimation\cite{47,51}.

This problem of high dimensional noisy data is compounded in cases where there is limited sensor data and consequently less number of features, especially in cases such as quantifying animal movements, cellular movement, particle diffusion or in diverse fields of science requiring non-invasive forms of measurements. As in such cases it becomes difficult to find reliable sensor data and data is gathered from a series of images such as satellite images to track animal migration or microscopic images to track cellular movements. Although PDEs are analysed to understand and quantify various ecological systems, the difficulty to model or paramaterize PDEs or form differential equations from data still remains. Various attempts have been made to to form differential models from a given data using likelihood or probability density functions\cite{39,40}
, but these techniques have become increasingly complicated and time consuming with multiple overlapping or interacting population and large datasets \cite{41}
. Reaction -diffusion equations have been used to model animal behaviors and understand biological pattern formation \cite{57,58}, however this approach still requires fitting the parameters to the PDEs.

Deep learning techniques have gained significant popularity in recent times partly due to their superior performance in various prediction and classification tasks\cite{13}
. Deep learning techniques use non-linear mapping to project input features and their interactions onto a output space thereby creating an approximated mathematical model for relating inputs to the outputs\cite{4,5}. 
Recent developments in deep learning to find data-drive solution of non-linear partial differential equations\cite{12,14}
has opened an avenue for extracting the learned differential equation from the network. Physics informed neural networks (PINN) can parameterize a PDE of known structure, which limits their application for PDE discovery from data with no prior knowledge of the system. Convolutional neural networks (CNNs) which are know for automatic feature discovery have been used to identify the structure of unknown equation\cite{48}, however parsimony of the equation variables is not guaranteed.

One of the major drawback of neural networks is that it is a black box model. Though there have been various attempts at explaining and interpreting deep learning systems\cite{8,9} there has not been an agreed upon explanation ]\cite{7} that links the weights of the neural networks and the function being approximated, since networks with different topology often provide same results. Furthermore, the obtained model is hidden, meaning it can only be viewed in terms of inputs and outputs which fails to provide any specific insights or physical interpretation of the function being approximated. This makes it extremely difficult to identify and extract specific function or equation being approximated by the network. These drawbacks of current techniques for model discovery demands for an all round approach which not only can handle noise, is easy to implement but also can provide interpretable models representing the system.

\begin{figure}[h! t]
\includegraphics[width=\textwidth] {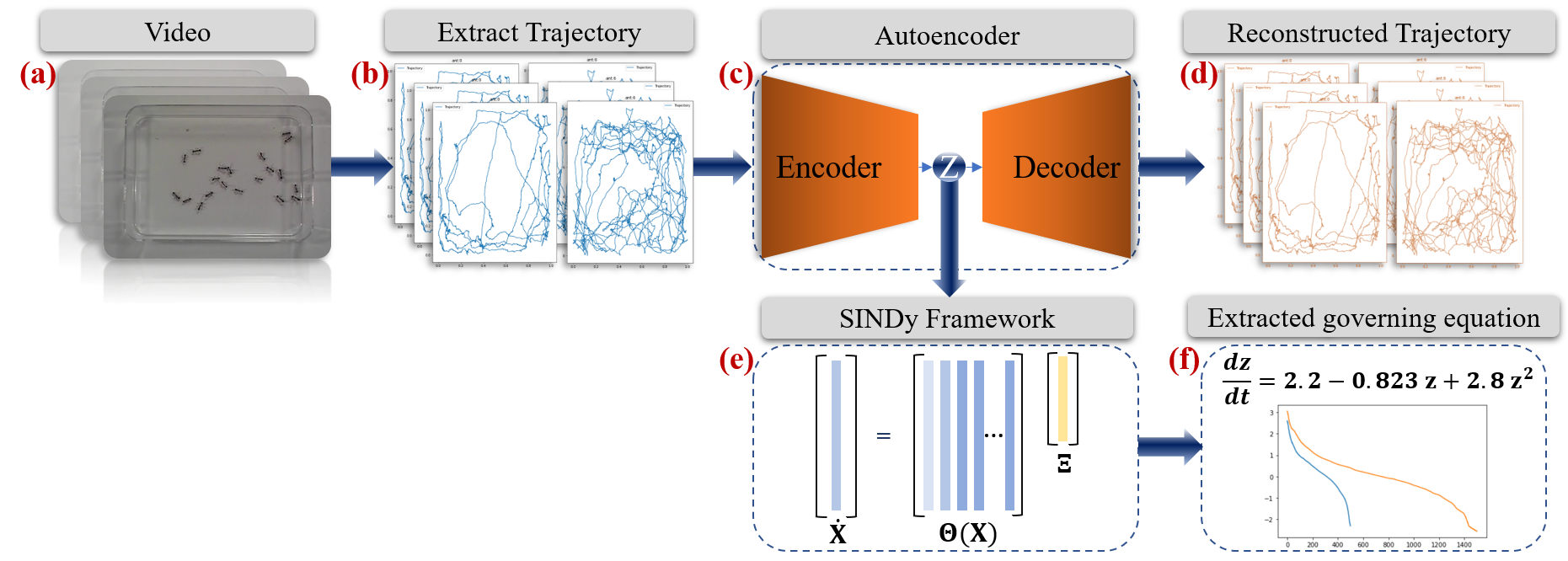}
\caption{Schematic of framework. Frame in videos (a) are used to extract states(trajectory of particle) of the system at each time step (b). The extracted data is then fed into LSTM variational autoencoder (c). The network is trained till the reconstructed states or trajectories match the input states (d) and using sparse regression (e) the latent representation of states at each time steps is extracted is then used to model time differential equation of the system (f). }
\label{fig: Fig1}
\end{figure}

Videos or sequence of images offer rich visual information about the dynamics of a system. Understanding scene dynamics is one of the core problems shared between computer vision and physical science communities. Physical parameters such as mass and friction\cite{59}, rigid body motion\cite{60} as well as Newtonian dynamics \cite{61} can be estimated by observing interactions of objects in a video. However, these approaches rely on prior knowledge of event dynamics and hence cant be extended to infer physics of a stochastic system.


In this paper, we present a deep learning based approach to obtain a time dependent differential equation representing the dynamics of multi particle/agent stochastic systems directly from videos. We extract the spatio-temporal trajectory data of individual particle/agent from videos of multi particle systems. The obtained system states are then projected onto a low dimensional latent space representing the states, this latent vector is used to form a parsimonious model using sparse regression based SINDy framework. Furthermore, we explore the ability of the obtained model to identify anomalous behaviour in the system. Figure\ref{fig: Fig1}. provides the general workflow of the framework where the input is a video of multi-particle system from which trajectories of particles is extracted using object tracking algorithm. LSTM based variational autoencoder is used to created embeddings of system sates and map the spatial features onto low dimensional latent vector. The extracted latent vector representing compressed states of the system is then distilled to find underling system dynamics.

\section{Methodology}
\subsection{Trajectory extraction:}
\begin{figure}[ht!]
\includegraphics[width=\textwidth] {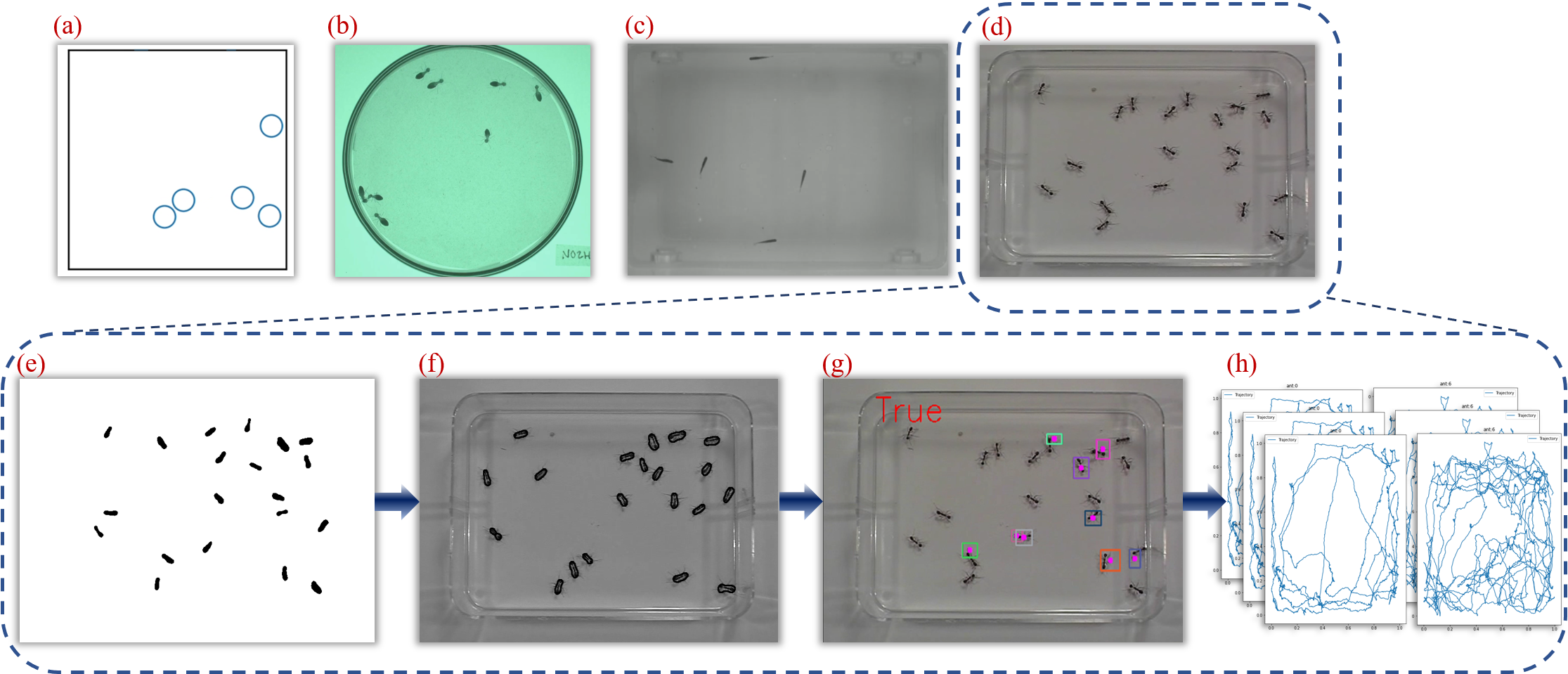}
\caption{ Examples of diverse multi agent system used in this study, (a) simulated multi particle system with elastic collision, (b) termites in confined space, (c) zebrafish behavior in box and (d) ants in a confined space. Trajectory extraction pipeline is demonstrated for ant system. Objects of interests are isolated based on color contrast and background removal (e). A contour is drawn around the object of interests (f) in each frame, the mid point of the drawn contour is then used to tag and track each object using CSRT algorithm\cite{30}.   }
\label{fig: Figure 2}
\end{figure}

In this study we explore our framework on diverse video datasets of 2 dimensional confined multi particle systems such as termite\cite{28}, zebrafish\cite{28} and ants in a box\cite{29}. We also analyze a system of simulated particles with elastic collision. Figure \ref{fig: Figure 2} (a),(b),(c),(d) shows various videos used in this study.

Tracking multiple objects or multi object tracking (MOT) in videos is an area of active research in the computer vision. Though there are various state of the art multi-object tracking systems available, since the priori of the all object appearance is known, all the particle are in-frame for the entire duration of the videos and there is no significant occlusion present, we have applied single object tracking model to solve MOT. In this study we use CSRT (Channel and Spatial Reliability Tracking) \cite{30} tracker available in OpenCV package to extract the trajectory of individual particle from the videos. CSRT tracker uses correlation filter with histogram of oriented gradients and colornames as features to tag objects, and searches the regions around last known position of the object in previous frames. In order to initialize, the tracker requires initial position of the objects. CSRT algorithm doesnot involve motion tracking i.e. it doesnot estimate the position of the objects in successive frames, this makes the output trajectories noisy but it also help to remove any algorithmic bias in the trajectory data, thus helping the us learn the 'true' dynamics of the system.

Figure \ref{fig: Figure 2} shows the general workflow for automatic trajectory extraction from ants in box video. First the video is cropped to a specific region of interest and using Gaussian blur, noise and variations in lighting and contrast is smoothed. For object detection in the first frame, a color and contrast threshold based detection algorithm is used. Isolating the object of interest in the first frame is done by using background subtraction and color isolation, as shown in Figure \ref{fig: Figure 2}e. Based on the this frame a contour is drawn around the objects of interests(\ref{fig: Figure 2}f) and it is tagged by drawing automated bounding boxes around the contours. The bounding box data is then used to initialize CSRT algorithm for tracking (\ref{fig: Figure 2}g). 

One limitation of this approach is when the objects are interact with each other the CSRT algorithm sometimes fails to distinguish the two particle, in such cases the size of bounding box was modified to enable it to find distinctive features on each object or the objects are tagged manually.

\subsection{Autoencoder: LSTM VAE}

Recurrent neural networks (RNN) are powerful deep learning algorithms designed to analyze sequential data and are preferred over multilayered perceptrons (MLP) to map a sequence onto a sequence.  The hidden states of RNNs are updated based on current as well as information from previous time steps, which make them appropriate to handle sequential time series data. LSTM\cite{15} and GRU\cite{16} are some of the most popular RNN architectures used to handle a variety of sequential data such as time series forecasting\cite{19}, forcasting chaotic systems \cite{27}, speech recognition\cite{20,21},translation\cite{22,23}, human dynamics\cite{26}. These architectures have shown to be promising compared to basic RNN structure which is difficult to train and has gradient vanishing problem\cite{17,18}. Moreover, LSTM networks have the ability to learn and reproduce long sequences and can handle spatio-temporal data like trajectories and have been extended to predict human trajectories in social settings\cite{24,25}. Auto encoders are typically unsupervised machine learning algorithms used to extract features, find compressed representation of the original data or to reconstruct and denoise data. LSTM autoencoders have be used in various fields\cite{33,34,35,36,37,38} especially to extract spatial-temporal features from videos. 

\begin{figure}[h!]
\includegraphics[width=\textwidth] {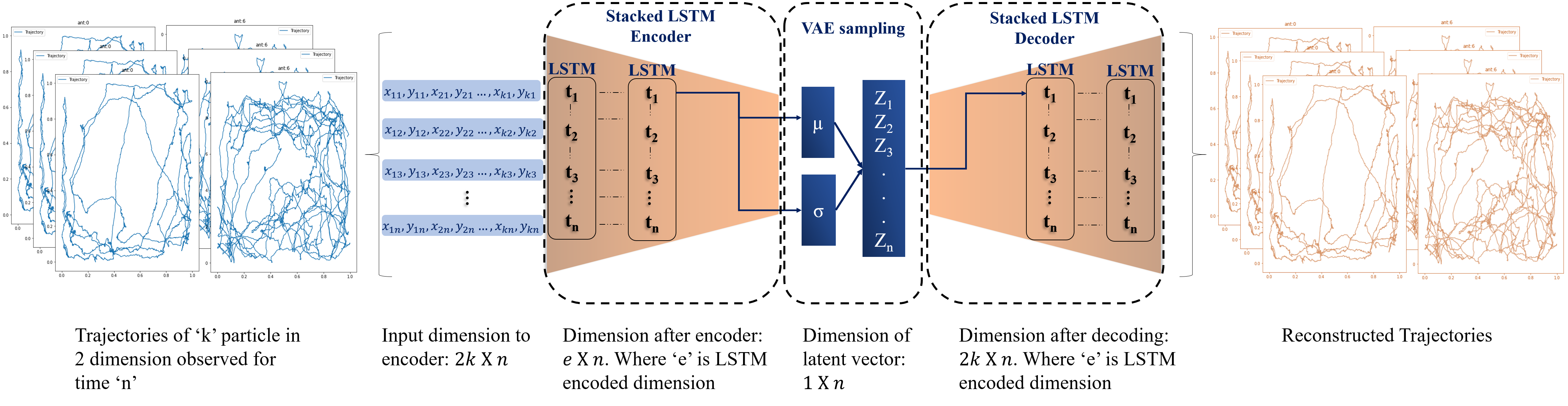}
\caption{Structure of LSTM Variational autoencoder. The position of particles at time 't' or the system state at each time step is used as an input to the variational autoencoder(VAE). The LSTM encoder projects the system states at each time step onto a embedding of size $eXn$. From this embedding a latent vector of size $1Xn$ is sampled. LSTM decoder then decodes this latent vector and reconstructs the input state.}
\label{fig:autoencderstructure}
\end{figure}

The extracted trajectory data provides with an time ordered spatial states of all particles in the system. Given the stochastic nature of the data and noise from tracking algorithm, the LSTM Variational autoencoder(VAE) encodes the state of the system at a given time step and learns a probabilistic latent distribution to ensure that the 'dominant' latent vector is sampled from the encoded latent distribution. 

LSTM VAE allows to map features at each time step to a sampled latent representation for the corresponding time step.  Figure \ref{fig:autoencderstructure}. shows the structure of LSTM-VAE. The trajectory data of each particle at time $'t'$ is in the given by $x_i^t,y_i^t$ where $i$ is the index of the particle. Hence ant system we have $'k'$ particles observed for $'n'$  time-steps will be a matrix of dimension  ( $2k$  $X $ $n$) , where we have $x_1,y_1,x_2,y_2,...x_{k},y_{k}$ features of k particles for n time-steps. These features are passed into the LSTM encoder, which maps these features to an encoded vector of predefined size. Here the feature extraction is done by LSTM encoder which finds features in spatial data at each time step.
The encoded vector from the encoder is further mapped to mean and standard deviation using a linear layer. During training a latent vector of size 1 is sampled from the encoder output. Hence, the $2k$ input features are mapped to a latent vector of size 1 for each time-step. For decoding, the latent vector is passed through a linear layer to obtain initial state for decoder which projects the latent vector to the output size which is same as that of input.
Since the system analysed in this study are stochastic and non-deterministic systems, more weight is given to the reconstruction loss which ensures that the reconstructed output is similar to the inputs i.e. the VAE is able to encode the trajectory of the particle to a single latent vector and that latent vector can reconstruct the original trajectories.  To ensure the best reconstruction and subsequently the best latent representation of the features, MSE loss and SmoothL1 loss is used depending on which loss function gives best reconstruction. 

\subsection{SINDy Framework}

The projected one dimensional latent space obtained from LSTM-VAE is considered to be representative of the spatial states of the system at corresponding time steps, leveraging this compressed representation we can describe the dynamics of the system in form of a time dependent differential equation. Sparse regression based approaches are the most commonly used to fit differential equation to measured data. In this work we have used SINDy framework which uses sparse regression to infer nonlinear dynamical systems from data. 

It is assumed that the systems can be described by the equation of form:
\begin{equation}
\frac{dz}{dt} = f(z)
\end{equation}
where, z is the obtained latent vector.

As per SINDy framework, $z$ is known and time derivative of z is calculated from data. Since, SINDy framework becomes unstable with noise, the latent vector was smoothed using Savitzky–Golay filter. A candidate library of size $n$ is constructed from the data as given by:
\begin{equation}
\Theta(z) = [1 \quad  z \quad  z^2\quad z^3\quad ...\quad z^n]
\end{equation}
A parsimonious model that includes most prominent and contributing terms is found by using sparse regression to solve:

\begin{equation}
\frac{dz}{dt} = \Theta(z) * \Xi
\end{equation}
where $\Xi$ is the library of coefficient corresponding to terms in $\Theta(z)$ and is calculated by sparse regression.  

In this study to find parsimonious model we have utilized, PySINDY\cite{52} library, which is a sparse regression package.

\section{Results}
We demonstrate the workflow on diverse video sets of multi-particle, ranging from termites and ants in a box, to zebra fish. We also demonstrate the workflow on simulated 2d multi particle system with elastic collision particle for interaction.

\subsection{Ants in a box}

The video shows 20 ants in a container and all ants are in frame throughout the video hence we considered all 20 ants in this study. Machine learning algorithms perform poorly for unscaled features. In previous works data was simulated from known PDEs and hence did not demand transformation or scaling of the features. Since, the scaling transformation impacts the identified differential equation in this study, a simple coordinated transformation technique provided in scikit-learn module\cite{53} is used which scales the features in range from 0 to 1. Figure\ref{ref:ant_traj} shows the scaled trajectories of 10 selected ants extracted from the video.

\begin{figure}[h!]
\centering
\begin{subfigure}{\textwidth}
\includegraphics[width=\textwidth] {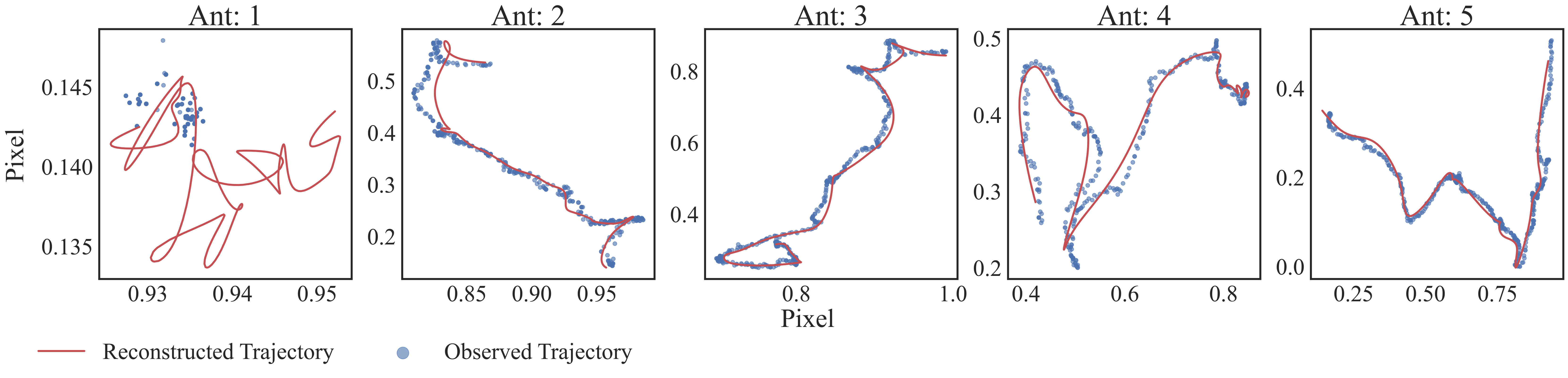}
\caption{Observed trajectories and reconstructed trajectories from LSTM VAE for 500 time steps. }
\end{subfigure}
\hfill
\hfill

\begin{subfigure}{\textwidth}
\centering

\includegraphics[width=0.99\textwidth] {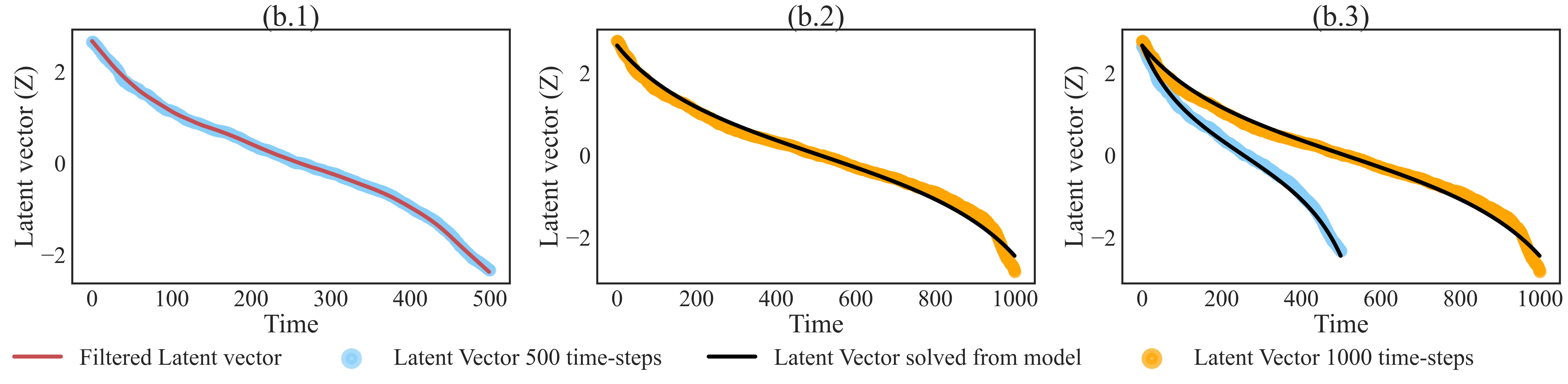}
\caption{Latent vector analysis obtained from LSTM VAE}

\end{subfigure}

\caption{Latent vector obtained by training VAE for 500 time steps is filter using Savitzky–Golay (a). Differential equation is modeled from the filtered latent vector using SINDy framework, the obtained equation is then solved for 500 time steps (b) and 1000 time steps(c). The latent vector solved from model for 1000 time steps shows good correlation with the latent vector obtained by training the VAE for 1000 time steps. }

\end{figure}

Each observed state of the system corresponds to 1 frame of the video. Since, the video has a frame rate of 30 frames/sec, 30 observations were taken for each second in the video. Ant system has 20 particles (ants) and considering 500 time-steps (frames) a matrix of dimension of $500X40$ is obtained, where  $x_1,y_1,x_2,y_2,...x_{20},y_{20}$  are the spatial features of 20 particles. These features are passed into stacked LSTM encoder with 2 LSTM layers, which maps these 40 features to a vector of size 32 for each time-step from which a latent vector of size 1 is obtained. Hence for 500 time steps a sequential latent vector $z$ of size $500X1$ is obtained.

Figure6 shows the reconstructed trajectories of 10 selected ants. It is observed that the neural network is able to approximate complex and stochastic trajectories of all ants. Based on these reconstructed trajectories we can say that latent vector obtained has the mappings of all the states of the system.  Better approximations can be obtained by a deeper network or by data augmentation and filtering. Furthermore, since these trajectories correspond to a short real-world time, it can be said that network has better approximation once the dynamics of each particle in the system is established (see appendix). 

To extract the underlining equation of the system represented by the latent vector, the latent vector '$z$' is first denoised using Savitzky–Golay filter of window size 51 and polynomial order 1.Figure 7. shows the latent vector and smoother latent vector. Following SINDy framework, we obtain the following equation representing the dynamics of the system up till 500 time steps:

\begin{equation}
\frac{dz}{dt} = -1.232z^2-3.266
\end{equation}
In order to ascertain viability of the model obtained in Eq. 4  to make describe current system states and predict future states, the model is solved using ODEINT python package for 500 (current) time steps and 1000 (future) time steps. To compare the predicted latent vector obtained from our model, `ground truth' latent vector is obtained by training the network for 1000 time steps. It can be observed from Figure 9. that the predicted latent vector obtained from solving the model defined by Eq. 4 shows good correlation with the ground truth latent vector.


\subsection{Zebrafish}






\begin{figure}[h!]
\centering
\begin{subfigure}[b]{0.45\textwidth}
\includegraphics[width=\textwidth]{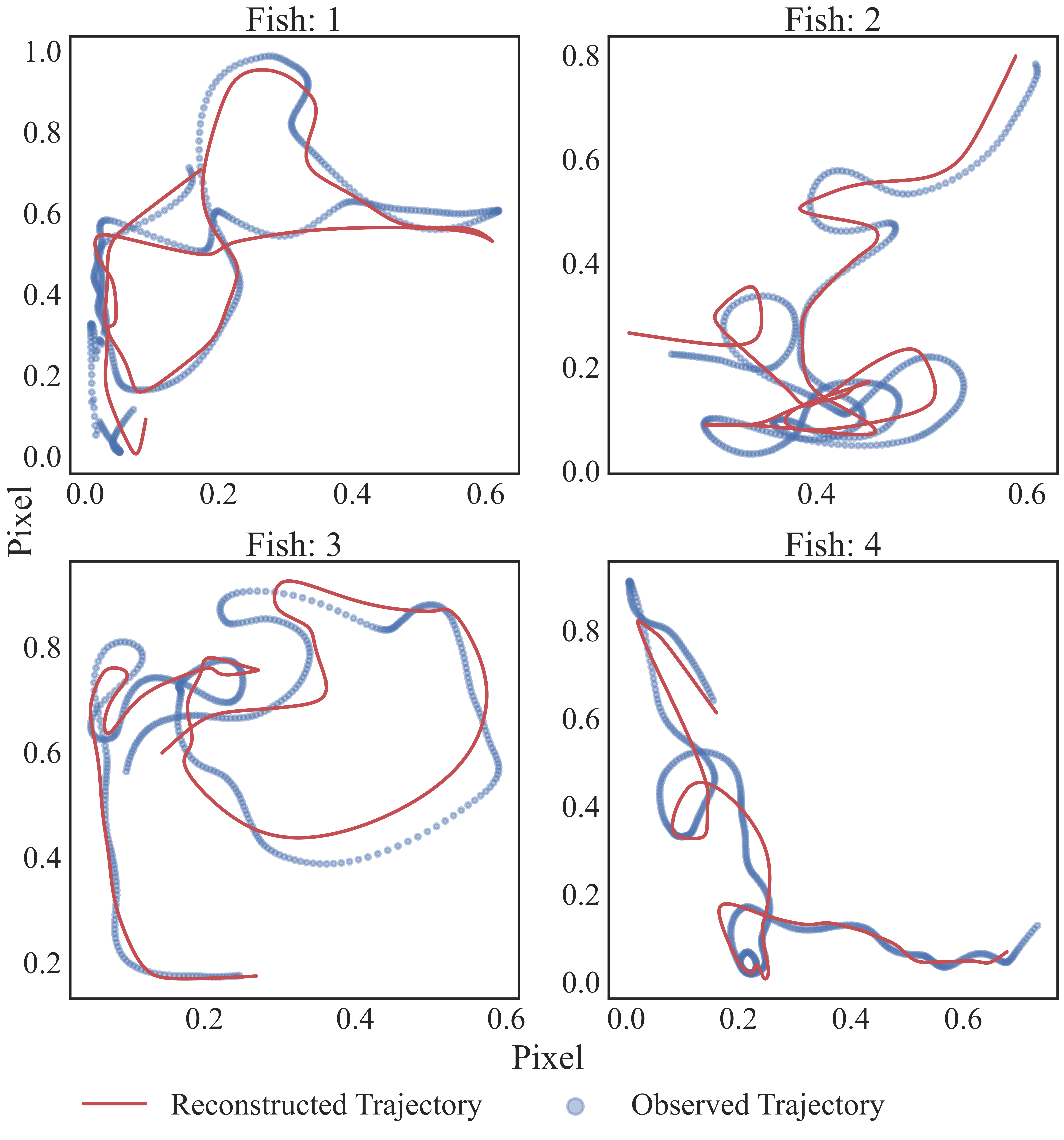}
\caption{Reconstructed and observed trajectories}
\end{subfigure}
\hfill
\begin{subfigure}[b]{0.45\textwidth}
\centering
\includegraphics[width=\textwidth]{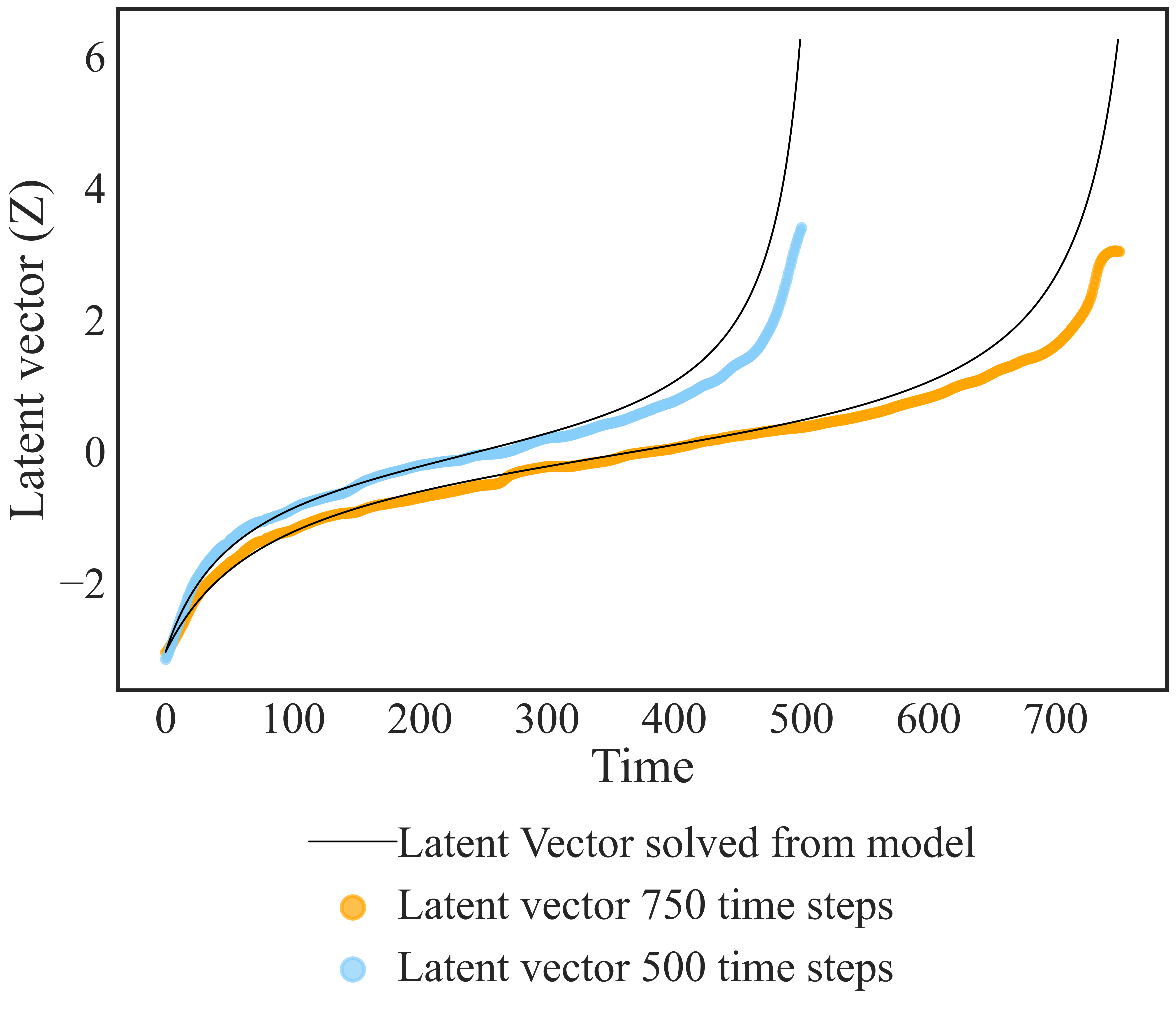}
\caption{Latent vector analysis}
\end{subfigure}
\caption{Dominant motion analysis on Zebrafish video dataset. Observed and reconstructed trajectories(a) show that the neural network is able to learn the dynamics of the system. The extrapolated latent vector model solved for 750 time steps shows good correlation with the latent vector latent vector obtained from training to 750 time steps (b).}
\label{fig:zeb}

\end{figure}

The zebrafish video set shows 5 fishes in a rectangular container and all the fishes are in frame for the entire duration of the video.  Since the data obtained from videos had significant noise, the trajectories were filtered using Savitzky–Golay filter of window size 31 and polynomial order 2. 
Similar to ant dataset, the x and y spatial coordinates of 5 fishes form 10 features defining the state of the system at any given time. These features are passed through one layer of LSTM encoder, where these 10 features are mapped to a vector of size 64 and from this encoded vector 1 latent vector is sampled for corresponding time step.

Obtained latent vector for 500 time steps was filtered similar to technique used in ant data set and using SINDy following equation is obtained:



\begin{equation*}
\frac{dz}{dt}=2.441 + 0.353 z + 3.020 z^2
\end{equation*}

The obtained model was validate by solving for 500 (current) time steps and for 750 (future) time steps. Comparing with the ground truth, as shown in Figure \ref{fig:zeb}. it can be observed that the model can predict the latent state.

\begin{figure}[h!]
\centering
\includegraphics[width=\textwidth]{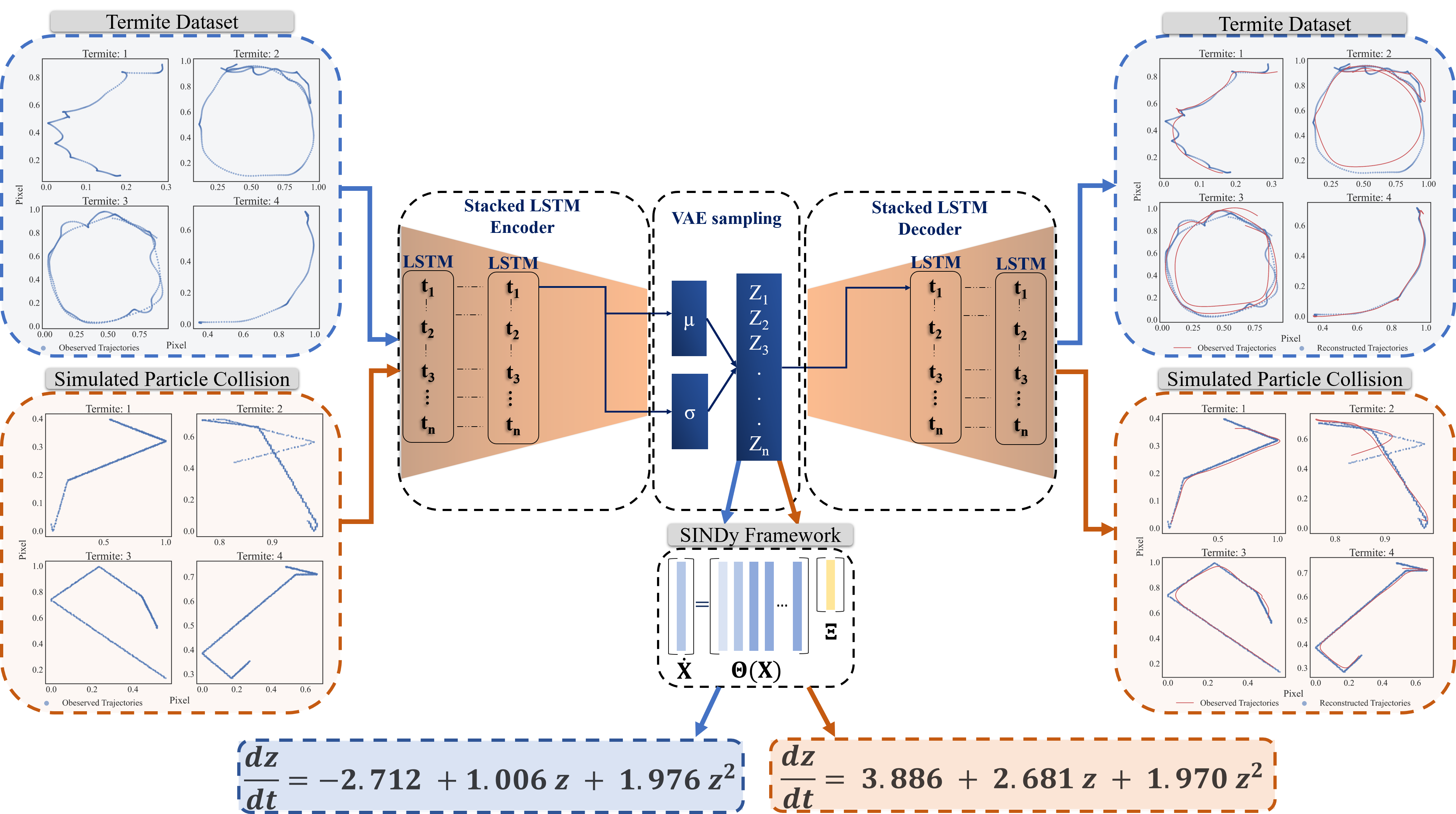}
\caption{Dominant motion analysis for Termite and collision dataset. }\label{fig:2data}
\end{figure}

Similar analysis was done for termite dataset and simulated elastic collision, the results are tabulate in Figure\ref{fig:2data}. (For details see SI:\ref{si:elas} and SI:\ref{si:termi})

\section{Discussion}
\subsection{Analysis of latent vector}
Latent vectors across these diverse systems show surprising similarities as long as the reconstructed trajectories are approximate of the true trajectories. Figure below shows comparison of latent vector of all data set. Since the latent vector follows similar trend, the governing equation derived from SINDy framework also provides equations of a common structure. Based on this common structure of the extracted governing equation,  it can be inferred that all theses diverse multi-particle systems follow a similar governing principle. 

To prove this and to find physical significance of latent vector, we need to look at long term trajectories of particles overtime. Figure below shows the 3D plot of trajectories of 2 randomly selected particles from all 4 dataset. It can be observed that all particles have a common trajectory behaviour, in which they start in the center of the frame and with time move towards to boundaries of the frame. Also no two particle (ants, termites and fishes in this case) can share same space at same time. The seemingly random behaviour when observed overtime is analogous to diffusion behaviour of fluid particles in a system in which the particles try to occupy the volume or the space available. In prior work animal movements are described by a class of PDEs called diffusion-taxis equations and methods methods have been developed to parametrize reaction-diffusion equations from animal location data\cite{56,54,55}.


\begin{figure}[h!]
\centering
\begin{subfigure}{\textwidth}
\includegraphics[width=\textwidth]{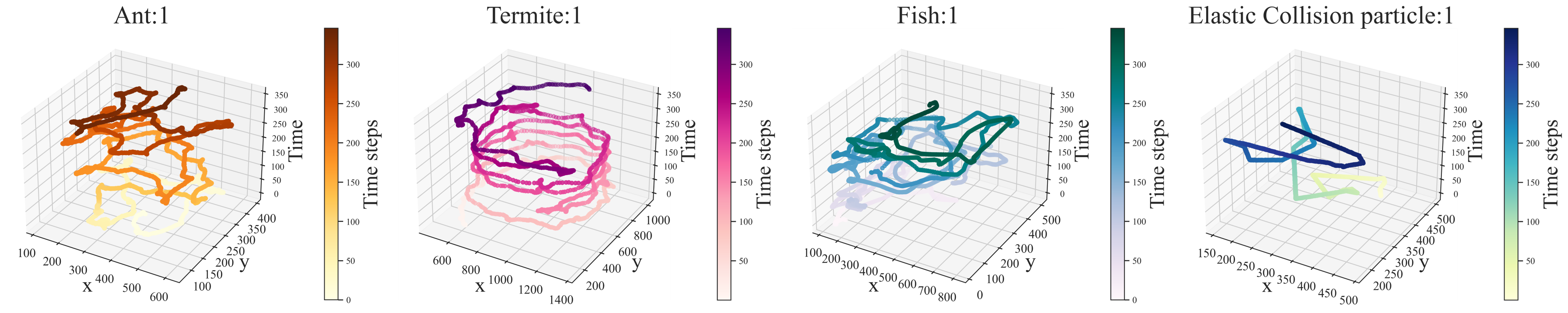}
\caption{Spatial trajectory of one particle through time}
\end{subfigure}
\hfill
\hfill
\begin{subfigure}{\textwidth}
\centering
\includegraphics[width=\textwidth]{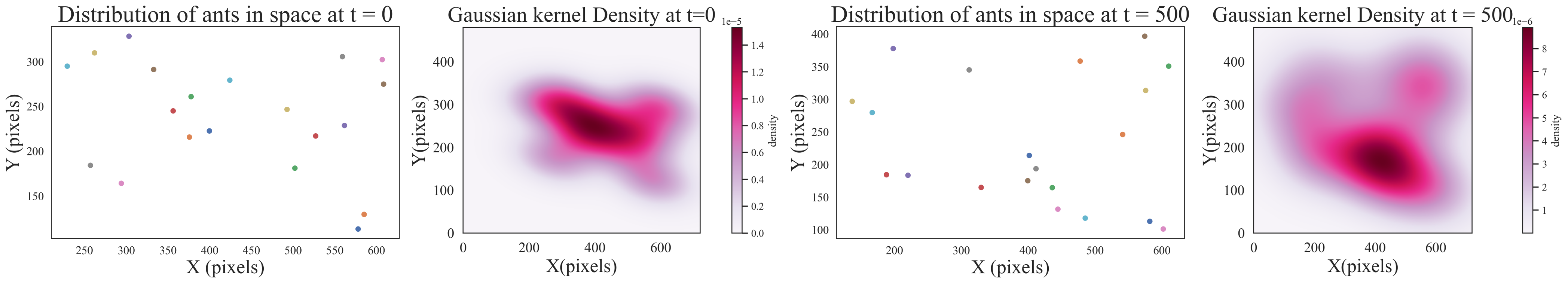}
\caption{Density distribution of ants at frame = 0 and frame = 500}
\end{subfigure}
\caption{It is observed that the particle in each of the system move towards the boundaries of the system (a). The density plots show that the probability of finding an ant in space in reducing with time i.e. the ants are 'dispersing' and occupying unexplored space in the system.}

\end{figure}

Furthermore, this 'diffusive' behaviour of the particle becomes more prominent when we observer variation in probability density of these particle overtime. Figure 12 shows density distribution  of ants calculated using Gaussian kernel estimation, at $time step=0$ and $time step=500$. It can be observed that the probability of finding ants at the center of the frame at $t=0$ is more compared to at $t=500$. Since the behaviour of ants is stochastic, comments can only be made based on the observed timesteps, hence we can say that the ants have diffused or occupied more space at $t=500$ compared to $t=0$.

The neural network is approximating the states of the system at each time step by creating non-linear mappings of state to the latent vector.  Based on the LSTM equation (See SI:\ref{si:lstm}) the hidden state at time `t' $(h_t)$ can be written as a function of $x_{t}$ and $x_{t-1}$ i.e.

\begin{equation}
h_t=f(x_t,x_{t-1})
\end{equation}

Thus, it can be inferred that the latent vector learned by the neural network is a function of change in spatial states of the system, in other words the encoder is embedding the state of the system to a function of average displacement, which is analogous to mean square displacement function in diffusion. This makes our network/algorithm highly useful to study transport phenomena, molecular dynamics, etc. Furthermore, given the stochastic nature of the data used opens up the possibility to use this network to find governing dynamics of a Brownian motion system. 

\subsection{Potential applications of obtained governing equation}
\begin{figure}[h!]
\centering
\includegraphics[width=\textwidth]{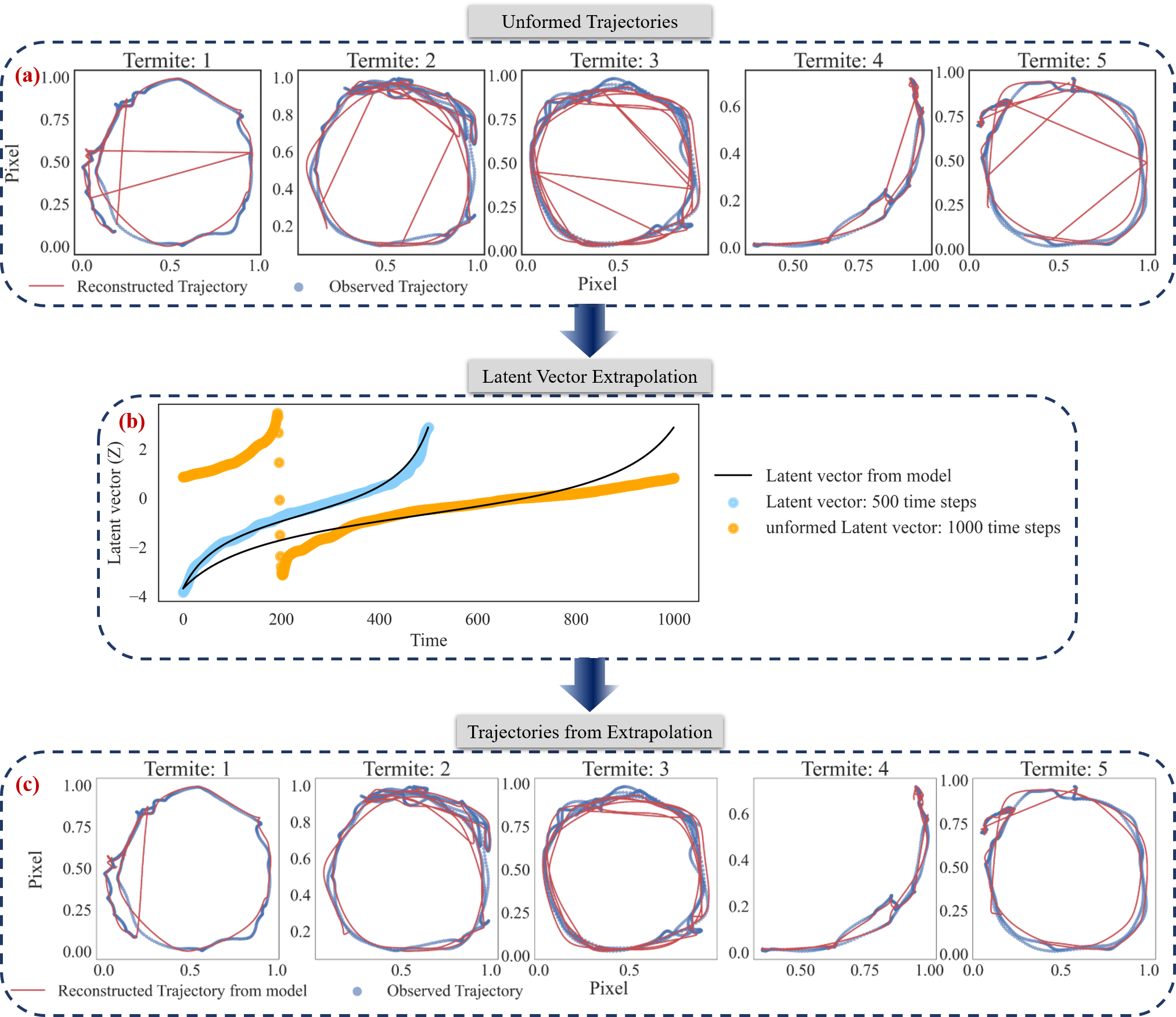}

\caption{To demonstrate that the model can be used to reconstruct and extrapolate unformed trajectories first we initialize the LSTM VAE for 1000 time steps and stop it prematurely before the states are learned (a). The latent vector of unformed states show significant deviation from the latent vector of fully formed states (b). The latent vector model obtained from training the network 500 time step is solved for 1000 time steps. The extrapolated latent vector is then fed into the decoder to form fully reconstructed states for 1000 time steps(c). }

\end{figure}

One of the potential application of using the latent vector and consequently obtained governing equation is for anomaly detection. The nature of the latent vector doesn't change with time (see SI:\ref{si:antlsall}) unless there is a disturbance or sudden change to the system. Thus the obtained equation can be used to detect anomaly if there are significant changes in the system. Furthermore, it is also observed that the latent vector follows the trend defined by the model as long as the reconstructed trajectory is similar to original trajectory. Thus, by changing the latent vector and by decoding the modified latent vector, different possible states of the system or simulate anomaly in the system can be simulated.

To prove both of these assertions, we trained the our network on termite dataset for 500 time steps and using SINDy framework extracted the differential equation representing state of the system from $t=0$ to $t=500$  from the latent vector.
To simulate anomalous behaviour,  we trained our network on same data set for 1000 time steps, this time stopping it prematurely to so that the reconstructed trajectories are not full formed. Figure\ref{fig:unforned_ls} shows the latent vector for 500 time steps and unformed latent vector for 1000 time steps, along with the vector derived from the model. Looking at the reconstructed trajectories and original trajectories, it can be observed that there are significant mismatched states.

We solve the differential equation obtained from latent vector representing 500 time steps, for 1000 time steps. This new vector obtained by solving the equation for 1000 time steps is then passed into the decoder which was stopped prematurely. Based on the reconstructed trajectories, it can be observed in Figure\ref{fig:reconstrued_fromlatetnt} that the mismatched states have been significantly reduced. 

This proves that the vector obtained from the model can not only be used to represent normal system behaviour but can also be used to filter anomalous/disturbed conditions of the system and conversely be used to simulate anomaly. It should be noted that an encode-decoder model trained at 500 time steps cannot fully predict the future states when latent vector of $t>500$ is passed through the decoder, but a reconstructions from a unformed encode-decoder model can reformed by solving the equation derived from latent vector representing previous system states.

\bibliographystyle{unsrt}  


\newpage
\setcounter{section}{0}

\part{Supplementary information}

Neural network architecture can be found here: https://github.com/BaratiLab/LSTM-VAE-for-dominant-motion-extraction
\section*{SI 1: LSTM cell}
\label{si:lstm}


\begin{figure}[h!]
\centering
\includegraphics[width=0.75\textwidth]{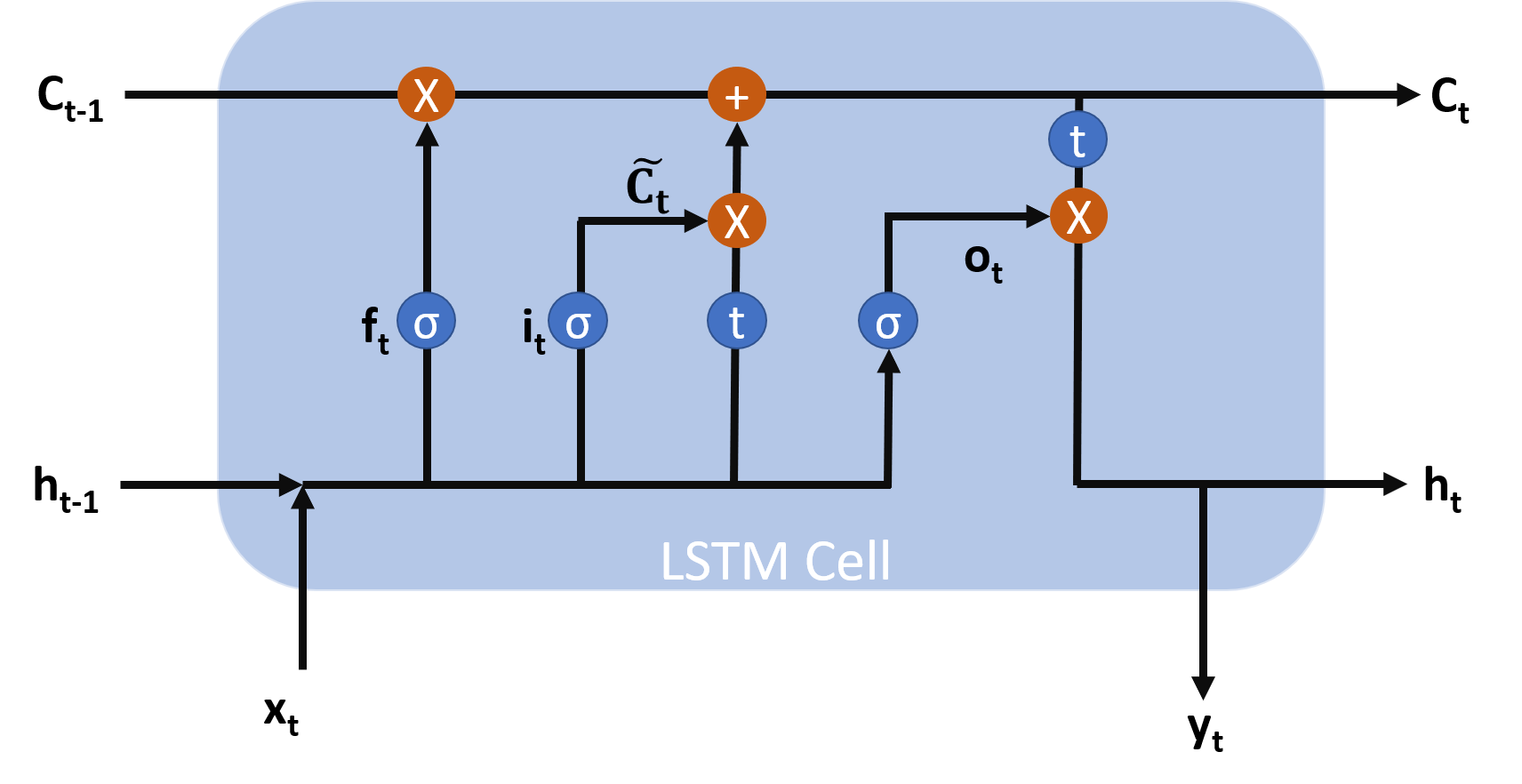}
\caption{Spatial trajectory of one particle through time}
\label{fig:lstmcell}
\end{figure}

The LSTM cell (Long short term memory) is a gated neural network, with 3 gates namely 1) input gate, 2)forget gate and 3) output gate. Figure\ref{fig:lstmcell} shows a typical LSTM cell. The gates of LSTM cell have sigmoid activation. The output from the cell and the hidden state of the cell depends on hidden state and cell state at previous time step as well as on the input state at current time step.

LSTM gates are governed by the following set of equations:

\begin{equation*}
\begin{aligned}
i_t =\sigma(w_i [h_{t-1},x_t]+b_i)
\\
f_t=\sigma(w_f[h_{t-1},x_t]+b_f)
\\o_t=\sigma(w_o[h_{t-1},x_t]+b_o)
\end{aligned}
\end{equation*}

Where, $i_t, f_t, o_t$ is the input, forget and output gates respectively. $x_t$ is the input at time `t' , $h_{t-1}$ is hidden state at time `t-1' and $w,b$ are the weights and bias of respective gates.

Based on the these gates, the outputs from the cell are:

\begin{equation*}
\begin{aligned}
C_t=f_t*c_{t-1}+i_t*tanh(w_c[h_{t-1},x_t]+b_c)
\\
h_t=o_t*tanh(C_t)
\end{aligned}
\end{equation*}
Here, $C_t$ is the cell state and $h_t$ is the hidden state. 

\section*{SI 2: Simulated Elastic collision system}
\label{si:elas}
\begin{figure}[h!]
\centering
\includegraphics[scale=0.5] {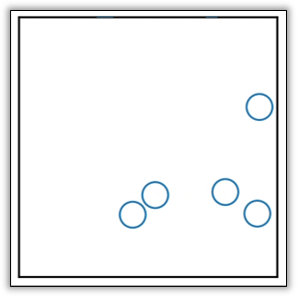}
\caption{Simulated elastic collision video frame.}
\label{fig:coll}
\end{figure}
This system is a physical simulation of two-dimensional interaction of particles in a confined space. In this paper we have used 5 particles of same size which undergoes elastic collision when they interact with each other or with the walls. In case of the interaction the total kinetic energy remains unchanged and the momentum is conserved. 
Based on law of conservation of momentum:

\begin{equation*}
m_1 *u_1+m_2*u_2=m_1*v_1+m_2*v_2
\end{equation*}

where, $m$ is the mass, $u$ is initial velocity and $v$ is final velocity after the collision. In this simulation the initial velocity at the start is randomly chosen.
but since we have particle of same size therefore same mass the interaction of particle in this simulation is defined by:

\begin{equation*}
u_1+u_2=v_1+v_2 
\end{equation*}

In order to track each particle we are using CSRT algorithm, bounding boxes were created manually around each particle. Form this video we get a total of 10 spatial features (x , y coordinate for each particle) for each frame(times step). These 10 features are used as inputs to the LSTM VAE, where these features are compressed to a latent vector representation of size 1 for each corresponding time step. SINDy framework is then used to find the underlining equation represented by the latent vector.

\section*{SI 3: Termite dataset}
\label{si:termi}
The termite dataset consist of 8 termites enclosed in a circular petri-dish as shown in Figure\ref{figure: Figure 2}b. Using CSRT tracker 64 spatial features (x,y coordinates) for each termite for each frame were captured. These 64 features were scaled in range 0-1 and then fed into LSTM VAE to obtain latent vector of size 1 for each time-step. The latent vector was smoothed using  Savitzky-Golay filter of window size 51 and order 1. Using SINDy framework on the filtered latent vector the system dynamics was found.

\section*{SI 4: Extracted Latent vector for multiple time steps}
\label{si:antlsall}
\begin{figure}[h!]
\centering
\includegraphics[width=0.75\textwidth] {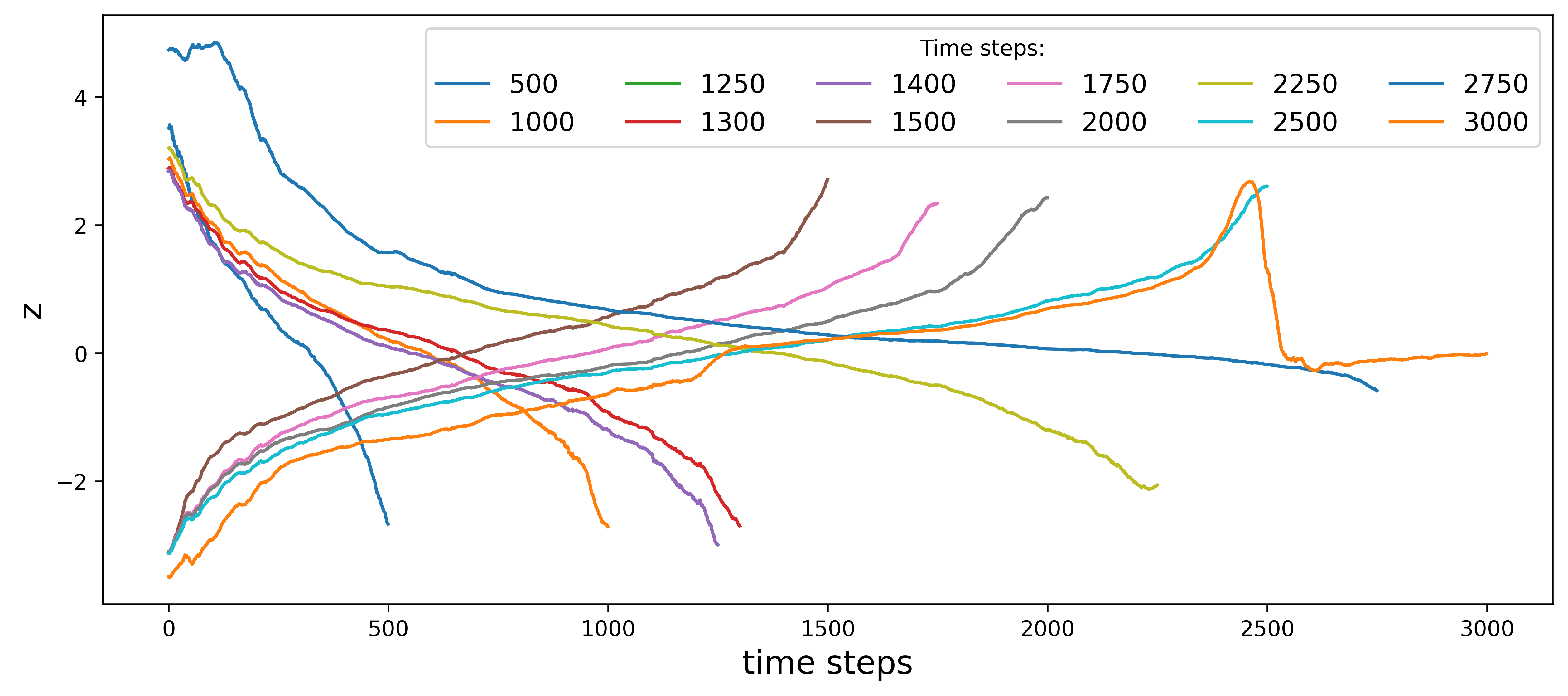}
\caption{Latent vector for ants in box dataset extracted for multiple time steps.}
\label{fig:ant mul}
\end{figure}
In order to ascertain the nature of latent vectors remains the same for far in future time, we trained LSTM VAE and extracted latent vector for various time steps. For this we used the ground truth trajectory provided with the ant video to compare and ensure that the nature of latent vector is same for the trajectory extracted using CSRT algorithm. The trajectory data has features for 10399 time steps. Figure\ref{fig:ant mul} show the latent vector for upto 3000 time steps. It can be observed that the long term dynamics of the system follows similar structure as the short term dynamics. The `flip' in the sign of latent vector observed after ~1400 time step has no affect on the system dynamics except for the change in signs of the coefficients. It was also observed that the even without the changing the signs of coefficients when dynamical system was solved for future time steps and fed into the trained decoder, the reconstructed trajectories were similar to the ground truth trajectories.

\begin{figure}[h!]
\centering
\includegraphics[scale=0.5] {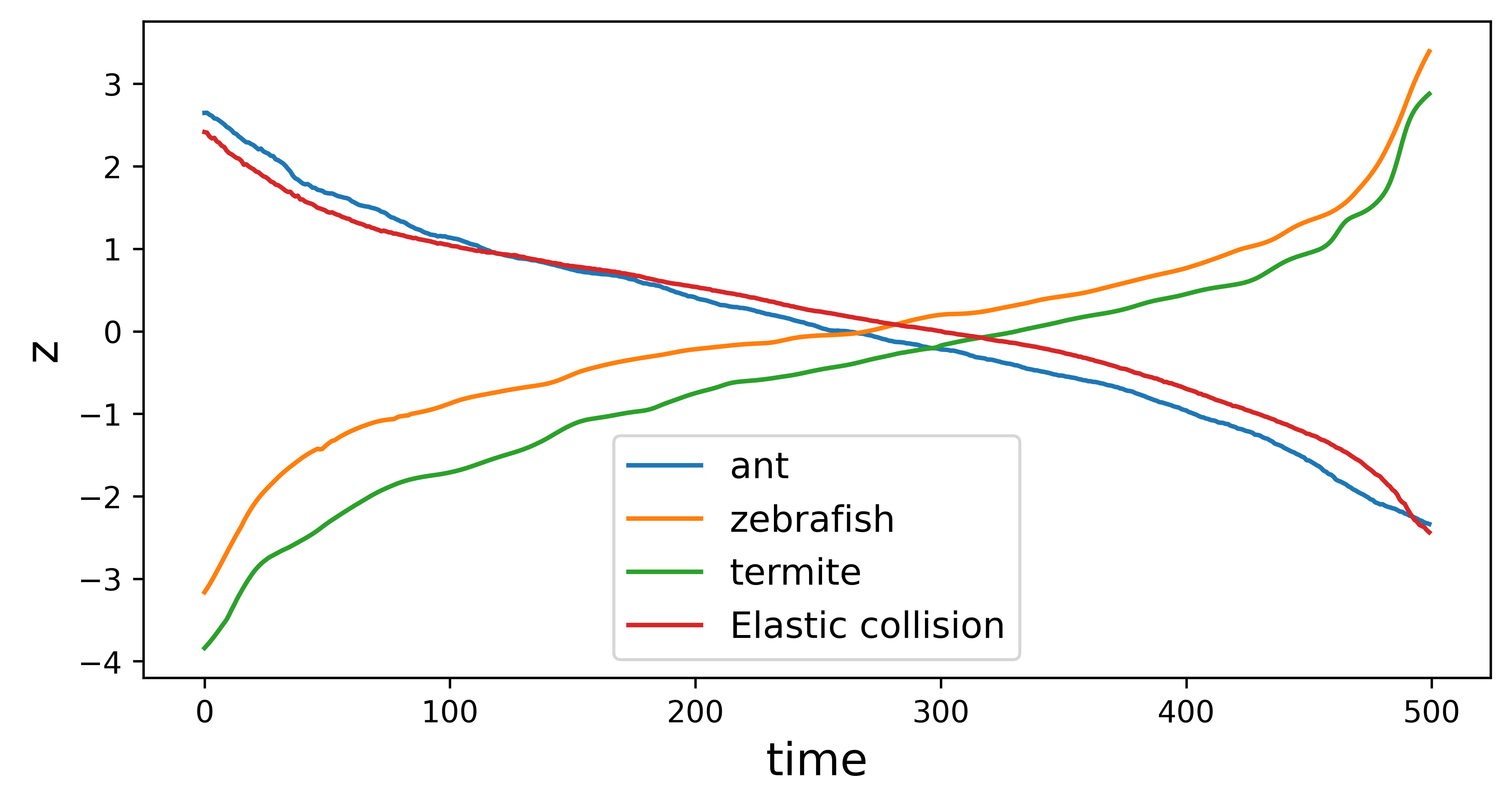}
\caption{Latent vector of 500 time steps for ant, termite, elastic collision and fish dataset.}
\label{fig:ant_diff}
\end{figure}

Figure\ref{fig:ant_diff} shows the similarity in the latent vectors obtained by training the network on different datasets.

\end{document}